\documentclass[10pt,twocolumn,letterpaper]{article}

\usepackage[pagenumbers]{cvpr} 

\usepackage{adjustbox}
\usepackage{array}

\newcolumntype{R}[2]{%
    >{\adjustbox{angle=#1,lap=\width-(#2)}\bgroup}%
    l%
    <{\egroup}%
}
\usepackage{graphicx}
\usepackage{amsmath}
\usepackage{amssymb}
\usepackage{booktabs}
\usepackage[lined,boxed,commentsnumbered,ruled]{algorithm2e}
\SetAlFnt{\small}
\SetAlCapFnt{\small}
\SetAlCapNameFnt{\small}
\SetAlCapHSkip{0pt}

\usepackage{float}
\usepackage{multirow}
\usepackage{color}

\usepackage[dvipsnames]{xcolor}

\definecolor{cvprblue}{rgb}{0.21,0.49,0.74}
\usepackage[pagebackref,breaklinks,colorlinks,citecolor=cvprblue]{hyperref}

\newcommand{\fakepara}[1]{\noindent\textbf{#1.}}

\title{Direct2.5$:$ Diverse Text-to-3D Generation via Multi-view 2.5D Diffusion}

\author{
Yuanxun Lu$^1$ \thanks{This project was performed during Yuanxun Lu's internship at Apple.} \quad Jingyang Zhang$^2$ \quad Shiwei Li$^2$ \quad Tian Fang$^2$ \quad David McKinnon$^2$ \\ Yanghai Tsin$^2$ \quad Long Quan$^3$ \quad Xun Cao$^1$ \quad Yao Yao$^1$ \thanks{Corresponding Author}
\and
$^1$Nanjing University\\
{\tt\small luyuanxun@smail.nju.edu.cn, \{caoxun,yaoyao\}@nju.edu.cn}
\and
$^2$Apple\\
{\tt\small \{jingyang\_zhang,shiwei,fangtian,dmckinnon,ytsin\}@apple.com}
\and
$^3$The Hong Kong University of Science and Technology\\
{\tt\small quan@cse.ust.hk}
}

\begin{document}

\maketitle

\begin{abstract}
Recent advances in generative AI have unveiled significant potential for the creation of 3D content. However, current methods either apply a pre-trained 2D diffusion model with the time-consuming score distillation sampling (SDS), or a direct 3D diffusion model trained on limited 3D data losing generation diversity. 
In this work, we approach the problem by employing a multi-view 2.5D diffusion fine-tuned from a pre-trained 2D diffusion model. The multi-view 2.5D diffusion directly models the structural distribution of 3D data, while still maintaining the strong generalization ability of the original 2D diffusion model, filling the gap between 2D diffusion-based and direct 3D diffusion-based methods for 3D content generation. 
During inference, multi-view normal maps are generated using the 2.5D diffusion, and a novel differentiable rasterization scheme is introduced to fuse the almost consistent multi-view normal maps into a consistent 3D model. We further design a normal-conditioned multi-view image generation module for fast appearance generation given the 3D geometry. Our method is a one-pass diffusion process and does not require any SDS optimization as post-processing. We demonstrate through extensive experiments that, our direct 2.5D generation with the specially-designed fusion scheme can achieve diverse, mode-seeking-free, and high-fidelity 3D content generation in only \textbf{10 seconds}. 
Project page: \href{https://nju-3dv.github.io/projects/direct25}{https://nju-3dv.github.io/projects/direct25}.

\end{abstract}

\section{Introduction}

Creating 3D content from generative models has become a heated research topic in the past year, which is key to a variety of downstream applications, including game and film industries, autonomous driving simulation, and virtual reality. Specifically, DreamFusion \cite{poole2022dreamfusion} was proposed to optimize a neural radiance field (NeRF)~\cite{mildenhall2021nerf} using a pre-trained 2D text-to-image diffusion model and the score distillation sampling (SDS) technique, showing promising results for text-to-3D generation of arbitrary objects without any 3D data. However, the indirect 3D probability distribution modeling inevitably deteriorates the final generation quality. For example, it has been reported in DreamFusion and its follow-ups \cite{wang2023score,lin2023magic3d,chen2023fantasia3d,wang2023prolificdreamer} that the overall generation success rate is low and the multi-face Janus problem exists.

Another line of work focuses on direct 3D generation by training on large-scale 3D data. For example, \cite{luo2021diffusion,nichol2022point} apply the probabilistic diffusion model for point cloud generation and ~\cite{Shim_2023_CVPR,jun2023shap} model the denoise diffusion process on signed distance field (SDF). These methods usually apply a specific 3D representation and train the denoise diffusion on such representation using a specific 3D dataset, e.g., ShapeNet~\cite{chang2015shapenet}, and show high-quality generation results on objects similar to the training set. However, the scale of the current 3D dataset is still too small when compared with the text-image data~\cite{schuhmann2022laion}. Even with the largest 3D dataset~\cite{deitke2023objaverse} available, it is still challenging to train a 3D diffusion model for diverse text-to-3D generation.

In this work, we instead extend existing text-to-2D models to a denoising diffusion process on multi-view 2.5D depth/normal data. Compared with full 3D representations such as 3D point clouds or meshes, 1) 2.5D information such as depth or normal are much easier to capture or collect (e.g., depth provided by active sensors); 2) the depth and normal maps perfectly align with the image data, making it possible to adapt and fine-tune a 2.5D model from a pre-trained 2D RGB model. In order to construct full 3D models, 2.5D maps viewed from multiple perspectives are necessary. Therefore, the target diffusion model should be capable of generating multi-view images with content consistency. In practice, we fine-tune existing text-to-image diffusion models on multi-view 2.5D renderings from the Objaverse dataset \cite{deitke2023objaverse}. On the one hand, the models are adapted to 2.5D information. On the other hand, joint multi-view distribution is captured with the help of structural modification of injecting multi-view information to the self-attention layers. During inference, multi-view images are generated synchronously by common schedulers like DDIM~\cite{song2020denoising}, which are then fused directly into a mesh by differentiable rasterization. The whole generation process completes in seconds, which is significantly faster than SDS-based methods that typically take 30 minutes. The system is extensively evaluated with complex text prompts and compared with both SDS-based and direct 3D generation methods, demonstrating the capability of generating 3D textured meshes with complex geometry, diversity, and high fidelity.

To summarize, major contributions of the paper include: 

\begin{itemize}[leftmargin=5mm]
    \item We propose to approach the 3D generation task by training a multi-view 2.5D diffusion model, which explicitly models the 3D geometry distribution while inheriting a strong generalization ability of the large-scale pre-trained 2D image diffusion. 
    \item We introduce an efficient differentiable rasterization scheme to optimize a textured mesh directly from the multi-view normal maps and RGB images. 
    \item We carefully design a generation pipeline that achieves diverse, mode-seeking-free, and high-fidelity 3D content generation in only 10 seconds. 
\end{itemize}

\begin{figure*}[t]
    \centering
    \includegraphics[width=\linewidth]{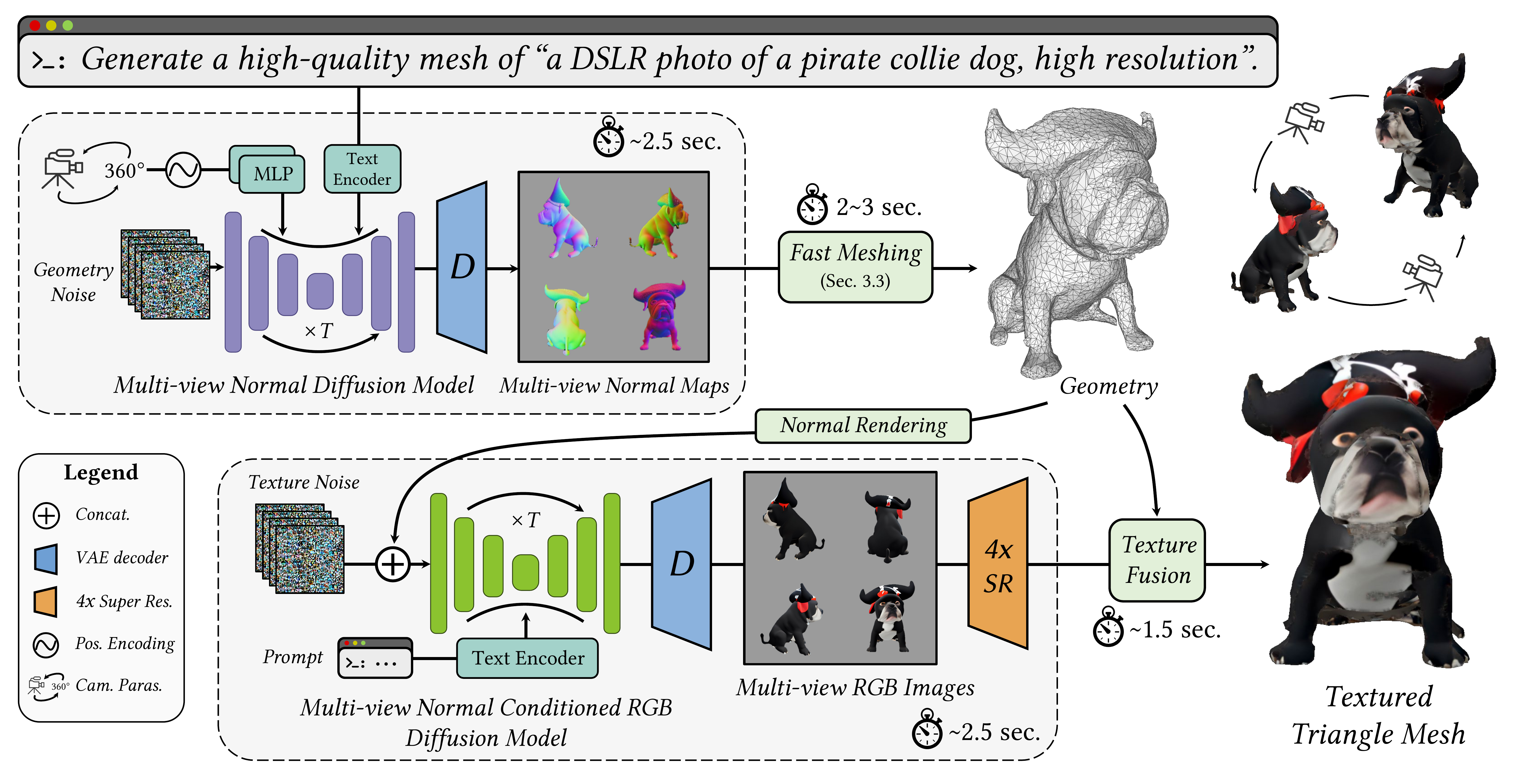}
    \vspace{-8mm}
    \caption{Overview of our text-to-3D content generation system. The generation is a two-stage process, first generating geoemtry and then appearance. Specifically, the system is composed of the following steps: 1) a single denoising process to simultaneously generate 4 normal maps; 2) fast mesh optimization by differentiable rasterization; 3) a single denoising process to generate 4 images conditioned on rendered normal maps; 4) texture construction from multi-view images. The whole generation process only takes 10 seconds.}
    \vspace{-4mm}
    \label{fig:method-system}
\end{figure*}

\section{Related Work}
\subsection{3D Generation by Score Distillation}
Score Distillation \cite{poole2022dreamfusion,wang2023score} is one of the most popular method recently for 3D Generation by pre-trained 2D diffusion models. It distillates the knowledge of image denoising to the optimization process of differentiable rendering systems so that randomly rendered views are gradually refined to describe the input text prompt. There are fundamental problems: 1) 2D diffusion models are not 3D-aware, and the generated samples have multi-face problem as a result; 2) Each optimization step requires single forward of the denoising UNet, making the whole process time consuming; 3) High guidance scale of prompts is preferred for better convergence, which leads to over-saturation of appearance; 4) the optimization is mode-seeking, losing the strong diversity of 2D diffusion model. Follow up works are proposed to solve some of them, but not all. Zero-1-to-3 \cite{liu2023zero} fine-tunes the 2D diffusion model with multi-view dataset to grant the ability of perspective control and mitigate the problem 1 in image-to-3D task. ProlificDreamer \cite{wang2023prolificdreamer} mitigate problem 3 and 4 by utilizing a KL-divergence loss to perform sampling instead of mode-seeking, at the cost of higher time complexity. In this work, we do not apply score distillation and completely separate diffusion process and 3D model optimization. The diffusion can be scheduled and conditioned normally, so that the results have diversity and realistic color. And the 3D model optimization operates on explicit representation so can be finished quickly. 

\subsection{Direct 3D Diffusion}
Fast 3D generation can be achieved by training a direct 3D diffusion model with 3D dataset. One key problem is to choose the 3D representation and design a special encoder/decoder for it. There are some early attempts to train direct 3D models for point cloud \cite{luo2021diffusion,nichol2022point,zhou20213d,zeng2022lion}, mesh \cite{liu2023meshdiffusion} and implicit representation like NeRF or SDF \cite{Shim_2023_CVPR,jun2023shap,gupta20233dgen,chen2023single}. However, they are trained on the limited datasets like ShapeNet \cite{chang2015shapenet} which have rather small data size, geometry complexity or category diversity. Recent 3D datasets such as Objaverse \cite{deitke2023objaverse} dramatically improve the state-of-the-art of 3D dataset, but is still limited compared to 2D image-caption datasets for training 2D diffusion models. In this work, we still use 2D neural network to deal with 2.5D maps, and thus we can perform fine-tuning on existing 2D diffusion models so as to inherit their strong generalization. 

\subsection{Multi-view Diffusion}
Generating multi-view images simultaneously is another strategy to bring 3D-awareness to 2D diffusion models. Two key modifications are proposed to achieve this: 1) Information from other views are concatenated with the current view as keys and queries in the self-attention layers. The gathered information can be from the single projection \cite{tang2023mvdiffusion}, epipolar lines \cite{tseng2023consistent, liu2023syncdreamer} or all the pixels \cite{shi2023mvdream}; 2) The model is fine-tuned on multi-view renderings from 3D dataset like Objaverse \cite{deitke2023objaverse}. To construct 3D models, previous works either use SDS \cite{shi2023mvdream}, which is still time-consuming, or image-based reconstruction systems like NeuS \cite{wang2021neus,liu2023syncdreamer, long2023wonder3d}, which requires at least 10 views to produce reasonable reconstructions. Similar to JointNet \cite{zhang2024jointnet} which explores the 2.5D domain, we choose to generate multi-view 2.5D maps like normal, so that we can use SDS-free reconstruction while still keep small view numbers.

\section{Method}
In this section, we introduce our multi-view 2.5D diffusion system, which synchronously generates multi-view 2.5D geometry images, i.e., normal maps, and corresponding texture maps given a text prompt as input for 3D content generation (Fig. \ref{fig:method-system}). Our method is efficient enough to generate various results in only 10 seconds. In Sec. \ref{sec:3.1}, we first briefly review the 2D diffusion model and formulate the multi-view 2.5D adaptation. We then illustrate the cross-view attention which enhances the multi-view consistency in Sec. \ref{sec:3.2}. In Sec. \ref{sec:3.3}, we describe how to produce the final 3D model from generated 2.5D geometry images, and finally in Sec. \ref{sec:3.4}, we demonstrate how to synthesize the texture maps given the generated normal maps, and construct the high-quality final textured triangle mesh.

\subsection{Diffusion Models and 2.5D Adaptation}
\label{sec:3.1}
Diffusion models learn a conversion from an isotropic Gaussian distribution to the target distribution (e.g. image spaces) via iterative denoising operations. We build our system on latent diffusion models (LDM), which contains a variational autoencoder (VAE) including an encoder and a decoder, a denoising network, and a condition input encoder. Compared to original diffusion models, LDM conducts the whole diffusion process in the latent image space and greatly improves efficiency and quality. Specifically, during the forward process, a noisy latent at time $t$ is sampled in the latent space and is gradually degraded by noise which makes it indistinguishable from the Gaussian noise, while the denoising process reverses the process, which iteratively predicts and remove the noise to get the real images. 

In this work, we extend 2D text-to-image diffusion models to generate multi-view geometry images. By fine-tuning a pre-trained 2D diffusion model using our 2.5D image dataset, we are able to inherit the generalization and also obtain the expressive generation ability for multi-view 2.5D geometry images. Let $(\boldsymbol{\mathcal{X}},c)$ be 3D data with caption from training dataset, $x_i \in \boldsymbol{\mathcal{X}}$ be multi-view renderings, $x_{i,t}$ be views corrupted by independent noise $\epsilon_i \in \boldsymbol{\mathcal{E}}$ at time $t$. The denoising neural network $\epsilon_{\theta}$ is trained by
\begin{align}
    L = \mathbb{E}_{(\boldsymbol{\mathcal{X}},c); \boldsymbol{\mathcal{E}} \sim N(0,1); t} \sum_{x_i \in \boldsymbol{\mathcal{X}}; \epsilon_i \in \boldsymbol{\mathcal{E}}} \| \epsilon_i - \epsilon_{\theta}(x_{i,t},c,t) \|_2^2.
\end{align}

\subsection{Cross-view Attention}
\label{sec:3.2}
Before fine-tuning, the multiple images generated from the base model for the same text prompt are not guaranteed to describe the same object because they are initiated from different noise maps and are denoised independently. We use a solution similar to \cite{shi2023mvdream}: we add data communication among the diffusion processes and fine-tune the model on multi-view image dataset to learn multi-view conditioning. Implementation-wise, we synchronize all the diffusion processes. When the calculation reaches a self-attention layer, we gather all the intermediate results as queries and values instead of just using the results from the current branch. Because images are treated as sequential inputs, the additional information can be simply concatenated together without introducing more trainable parameters. This architecture ensures that the diffusion processes are mutually conditioned, which serves as a structural prerequisite for multi-view consistent generation.

\subsection{Explicit Multi-view 2.5D Fusion}
\label{sec:3.3}
There are various approaches available for constructing a 3D model from multi-view observations. Among them, image-based 3D reconstruction methods such as multi-view stereo \cite{yao2018mvsnet, yao2019recurrent, zhang2023vis, gu2020cascade} or NeRF \cite{mildenhall2021nerf, muller2022instant, barron2021mip, fridovich2023k} requires at least 10 images for high-fidelity reconstruction, which pose significant computational challenges in multi-view diffusion scenarios. However, by taking benefits from 2.5D information, one could effectively reduce this requirement. In practice, we generate 4 normal maps aligned with world coordinates from different viewpoints (front, left, right, and back). To fuse these observations into a triangle mesh, we explore the insight of geometry optimization from an initialized mesh via differentiable rasterization. This optimization, which is independent of neural network inference, achieves convergence rapidly within seconds (see Alg.~\ref{alg:geometry_opt}).

\begin{algorithm}[t!]
\caption{\emph{Multi-view Geometry Optimization}}
\label{alg:geometry_opt}
\SetAlgoLined
\KwIn{Multi-view normal maps $I_i$ and camera parameters $\pi_i$, where $i \in \{0,1,2,3\}$}
\DontPrintSemicolon
\SetKwInOut{Parameters}{Parameters}
\KwOut{$M=(V,F)$ output triangle mesh\;}
\Parameters{}
\quad $T$: max number of optimization iterations\;
\quad $\lambda_{\alpha}, \lambda_{nc}$: weights for alpha and normal consistency loss \;
\BlankLine
$V_{occ} \leftarrow$ InitOccupancyVolume \\
\For{i $\in \{\textup{0, 1, 2, 3}\}$}{
    \textup{Compute alpha mask $\alpha_i$} $\leftarrow$ \textup{thresholding($I_i$)} \\
    \textup{Update $V_{occ}$ $\leftarrow$ SpaceCarving($\alpha_i, \pi_i$)}
}
\BlankLine
$M \leftarrow$ \textup{MarchingCubes($V_{occ}$)}\\
$M \leftarrow$ \textup{MeshSimplification}($M$)\\
\BlankLine
\For{iter $\leftarrow$ $T$}{
   $\hat{I}, \hat{\alpha} \leftarrow$ \textup{DifferentiableRender($M, \pi$)} \\
   $loss \leftarrow \mathcal{L}_{n}(I, \hat{I})$ + $\lambda_{\alpha} \mathcal{L}_{\alpha}(\alpha, \hat{\alpha})$ + $ \lambda_{nc} \mathcal{L}_{nc} (M)$ \\
   \textup{Optimize($loss$)} \\
   $M \leftarrow$ \textup{Remesh}($M$) 
}
\end{algorithm}

\begin{figure}[t]
    \centering
    \includegraphics[width=\linewidth]{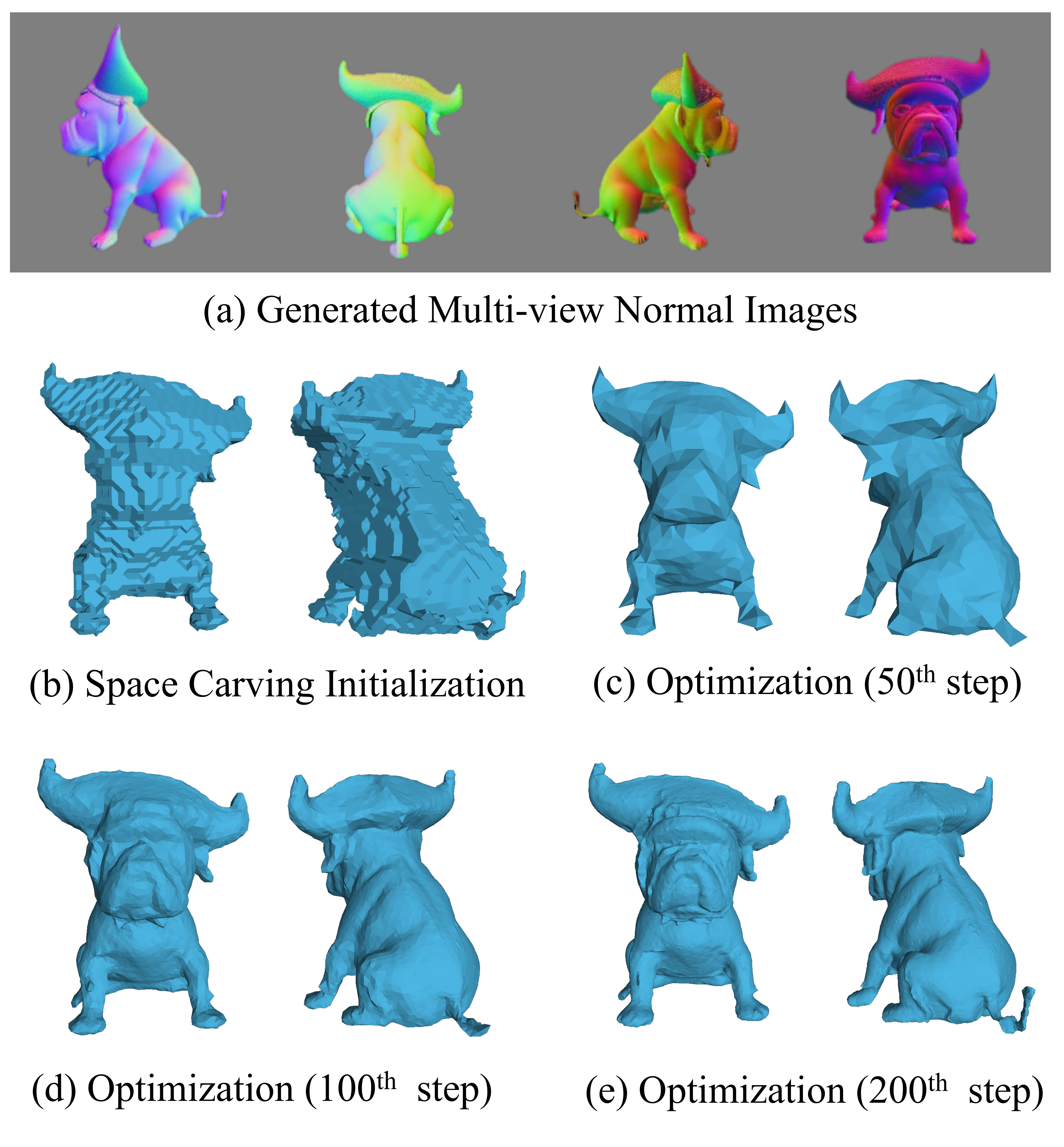}
    \vspace{-8mm}
    \caption{Illustration of explicit geometry optimization. (a) is the generated normal images given a prompt "a DSLR photo of a pirate collie dog, high resolution”. (b) shows the space carving initialization results mesh in the front and side views. (c), (d), (e) present the intermediate optimization states at 50, 100, 200 steps, separately. As shown, 200 steps are enough to reconstruct the fine details like the skin folds of the dog's face and the thin dog tail.}
    \vspace{-4mm}
    \label{fig:geo-opt}
\end{figure}

\noindent \textbf{Space Carving Initialization}. A simplistic and straightforward approach would be to initialize the shape using basic geometric primitives like spheres and cubes and optimize. However, this often introduces significant challenges during the latter geometry optimization, particularly when the target shape's topology diverges significantly from these elementary forms. To tackle this challenge, we employ the space carving algorithm \cite{kutulakos2000theory} for shape topology initialization. Besides, it also provides a good initialization for latter geometry optimization. Fig. \ref{fig:geo-opt} (a) shows the space carving results. Specifically, this process begins by segregating the background normal maps through a simple value thresholding. Subsequently, a volume in the interested space is created, and each voxel is projected onto the images using the camera parameters, determining whether the corresponding pixel is part of the object or the background. 
By gathering all projections under different views, we construct an occupancy volume, in which a voxel's occupancy is set to 0 (indicating emptiness) if all of its projections belong to the background, and 1 (indicating occupancy) otherwise. Finally, we apply the marching cubes \cite{lorensen1987marching} on the occupancy volume to extract the zero level-set surface to form the initialized shape. This technique not only effectively preserves the topology, but also provides a rough shape estimation generated from the multi-view normal images.

\begin{figure*}[t]
    \centering
    \includegraphics[width=\linewidth]{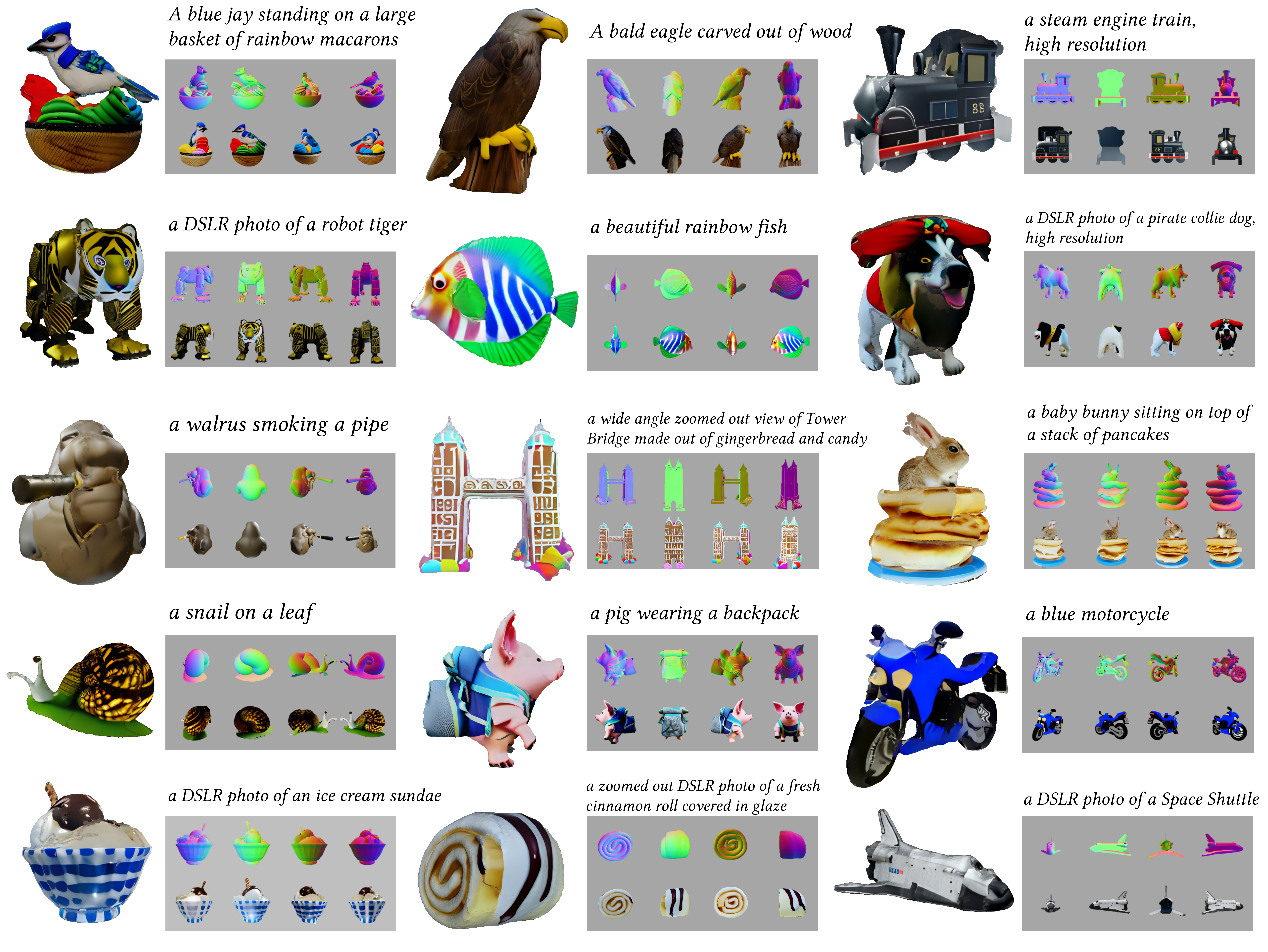}
    \vspace{-8mm}
    \caption{A gallery of our text-to-3d generation results. Given text prompts as description input, our method outputs high-quality textured triangle mesh in only 10 seconds. Note that the prompts are not from the training set. Best viewed zoomed in.}
    \vspace{-5mm}
    \label{fig:results_gallery}
\end{figure*}

\noindent \textbf{Optimization via Differentiable Rasterization}. Once we have obtained the initialized geometry, we further refine the mesh details based on observational data. This refinement is mathematically formulated as an optimization problem, targeting the triangle triangle vertices $V$ and faces $F$. As illustrated in Alg.~\ref{alg:geometry_opt} and Fig. \ref{fig:geo-opt}, we first simply the marching cube-generated mesh to a lower face number, which is found to help accelerate and improve the optimization. In each optimization step, we optimize the model by minimizing the $L_1$ loss between the rendered results and observations, as well as a normal consistency regularization. The loss function could be written as follows:

\begin{flalign}
     \mathcal{L}_{V}  = \mathcal{L}_{n} + \lambda_{\alpha} \mathcal{L}_{\alpha} + \lambda_{nc} \mathcal{L}_{nc},
\end{flalign}

where $\mathcal{L}_{n} = \frac{1}{4} \sum_i^4 || n_i - \hat{n_i} ||_1$ is the normal rendering loss. It measures the mean $L_1$ distance between rendered normal maps $n$ and the observations $\hat{n}$ under different camera viewpoints $i \in \{0, 1, 2, 3\}$. Similarly, $L_{\alpha} = \frac{1}{4} \sum_i^4 || \alpha_i - \hat{\alpha_i} ||_1$ is the alpha mask loss, which computes the difference between rasterized object mask $\alpha$ and the observed $\hat{\alpha}$, and the latter could be obtained by a simple value thresholding $\delta = 0.05$ in the generated normal maps. 

We additionally integrate a normal consistency term, denoted as $\mathcal{L}_{nc}$ to regularize the mesh. Specifically, this regularization is designed to smooth the mesh on a global scale by minimizing the negative cosine similarity between connected face normals. The hyperparameters $\lambda_{\alpha}, \lambda_{nc}$ which control the different weights for alpha mask loss and normal consistency regularization are set to 1 and 0.1 respectively. We adopt the nvdiffrast library \cite{laine2020modular} for differentiable rasterization.

After each optimization step, we further perform remeshing by merging or splitting triangle faces using the strategy from \cite{palfinger2022continuous}. During experiments, we empirically found that only about 200 optimization steps are enough to generate a high-quality geometry mesh, which takes only around 2 to 3 seconds. As shown in the fig. \ref{fig:geo-opt} (c-e), the dog shape has been well optimized at around 200 steps.

\begin{figure*}[t]
    \centering
    \includegraphics[width=\linewidth]{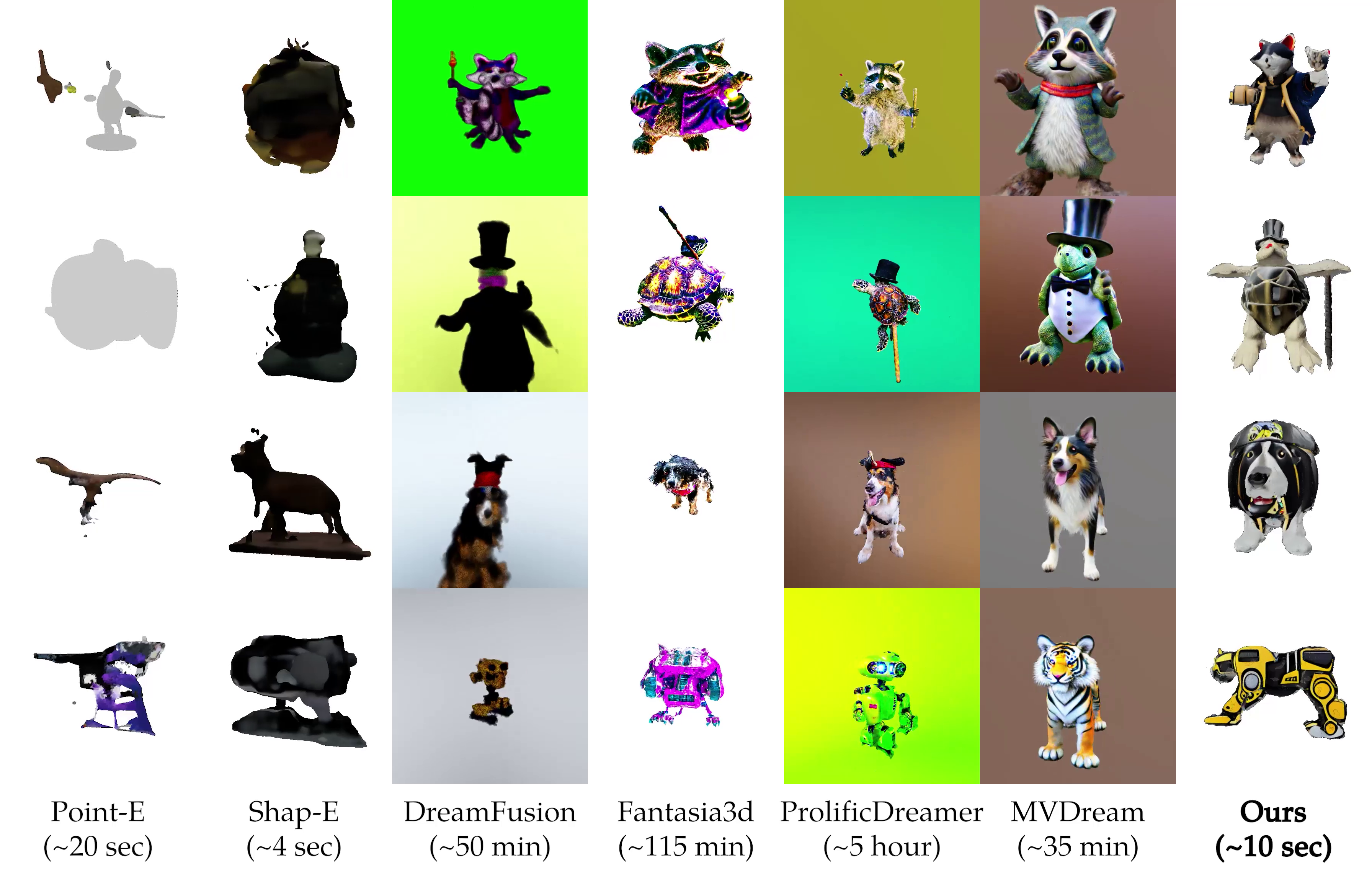}
    \vspace{-8mm}
    \caption{Qualitative comparisons. Direct 3D diffusion systems are not well generalized to the complex prompts. SDS-based methods except MVDream are slow and suffered from multi-face and over-saturation problems. MVDream can generate realistic geometry and appearance with fine details but has limited diversity (Fig.~\ref{fig:var-comp}). In contrast, our system can generate realistic 3D models efficiently. Input prompts: 1) a zoomed out DSLR photo of a wizard raccoon casting a spell, 2) a DSLR photo of a turtle standing on its hind legs, wearing a top hat and holding a cane, 3) a DSLR photo of a pirate collie dog, high resolution, and 4) a DSLR photo of a robot tiger.}
    \vspace{-4mm}
    \label{fig:qual-comp}
\end{figure*}

\subsection{Texture Synthesis}
\label{sec:3.4}
Texturing the mesh is another crucial step in achieving a high-quality result. Similar to the geometry generation, we initially synthesized multi-view texture maps, which were then applied to the generated geometry. In practice, another multi-view diffusion model generates the corresponding multi-view texture maps, conditioned on text prompts and the multi-view normal images.

As shown in figure \ref{fig:method-system}, the architecture of the multi-view normal-conditioned diffusion model is similar to the text-to-normal model, except that we extend the first convolution layer by increasing the number of channels to satisfy the normal latent condition input. Specifically, we initialize the extra trainable parameters in the first layer to zero before training. The normal condition plays a pivotal role in shape information and guides the model to generate both text- and shape-aligned texture images. We further apply super-resolution, i.e., Real-ESRGAN \cite{wang2021realesrgan} on the generated texture maps to increase more appearance details, resulting in a 4 $\times$ resolution upscale from $256\times256$ to $1024\times1024$.

After obtaining the high-resolution RGB images, the final stage is to project these images to the shape geometry and generate a global texture. We perform UV parameterization and the Poisson blending algorithm \cite{Waechter2014Texturing} to alleviate multi-view inconsistency.

\noindent \textbf{Iterative updating}. In most cases, a single run of the pipeline is enough to generate high-quality results. However, since we generate 4-view information at once, there may be some areas unobserved in the generated RGB images (such as the top area of the object), and a texture refinement is required. To address this issue, we could iteratively update the generated images by using popular inpainting \cite{lugmayr2022repaint} pipelines in diffusion models to refine the generated textures. By computing a visibility mask at a new camera viewpoint, the invisible areas could be generated given a certain noise strength. During experiments, we found that only 1 or 2 iterations are enough to inpaint the unseen areas.

\section{Implementation Details}
In the following, we describe the aspects relevant to our system implementation details: dataset preparation in Sec. \ref{sec:4.1} and training setups in Sec. \ref{sec:4.2}.

\subsection{Dataset Preparation}
\label{sec:4.1}
We use the Objaverse~\cite{deitke2023objaverse} dataset for 2.5D training data generation, which is a large-scale 3D object dataset containing 800K high-quality models. We use the captions provided by cap3d \cite{luo2023scalable} as text prompts. We filter the dataset by sorting the CLIP scores and selecting the top 500K objects with high text-image consistency. Each object is firstly normalized at the center, and we render the scene from 32 viewpoints uniformly distributed in azimuth angles.

Besides, we also adopt a large-scale 2D image-text dataset to improve the generation diversity. Specifically, we use the COYO-700M dataset \cite{kakaobrain2022coyo-700m}, which also contains metadata like resolution and CLIP scores \cite{radford2021learning}. We filter the dataset with both width and height greater than 512, aesthetic scores \cite{schuhmann2022improved} greater than 5, and watermark scores lower than 0.5, which results in a 65M-size subset. Though the filtered dataset is reduced to 1/10 of the original size, it is still larger than the 3D dataset. Actually, we do not use the whole filtered dataset during training. 

Please check the supplementary for more details.

\subsection{Training Setup}
\label{sec:4.2}
As introduced above, we train the model with both 2.5D rendered images and natural images, with a probability of 80\% to select the former. This makes the instances seen in each batch nearly equal for two kinds of data. We use the Stable Diffusion v2.1 base model as our backbone model and fine-tune the latent UNet only for another 50K steps with 1000 warmup steps. Similar to Zero123 \cite{liu2023zero}, we use an image sample size of $256 \times 256$ for better and faster training convergence. The learning rate is set to $1e-5$. We drop the text prompt conditioning with a probability of 15\% and apply a noise offset of 0.05. The full training procedure is conducted on 32 NVIDIA A100 80G GPUs (800K steps for the text-to-normal model and 18K steps for the normal-conditioned RGB model, which takes around 80 and 20 hours separately). The batch size is set to 45 on each GPU which leads to a total batch size of 1440.

\begin{figure}[t]
    \centering
    \includegraphics[width=\linewidth]{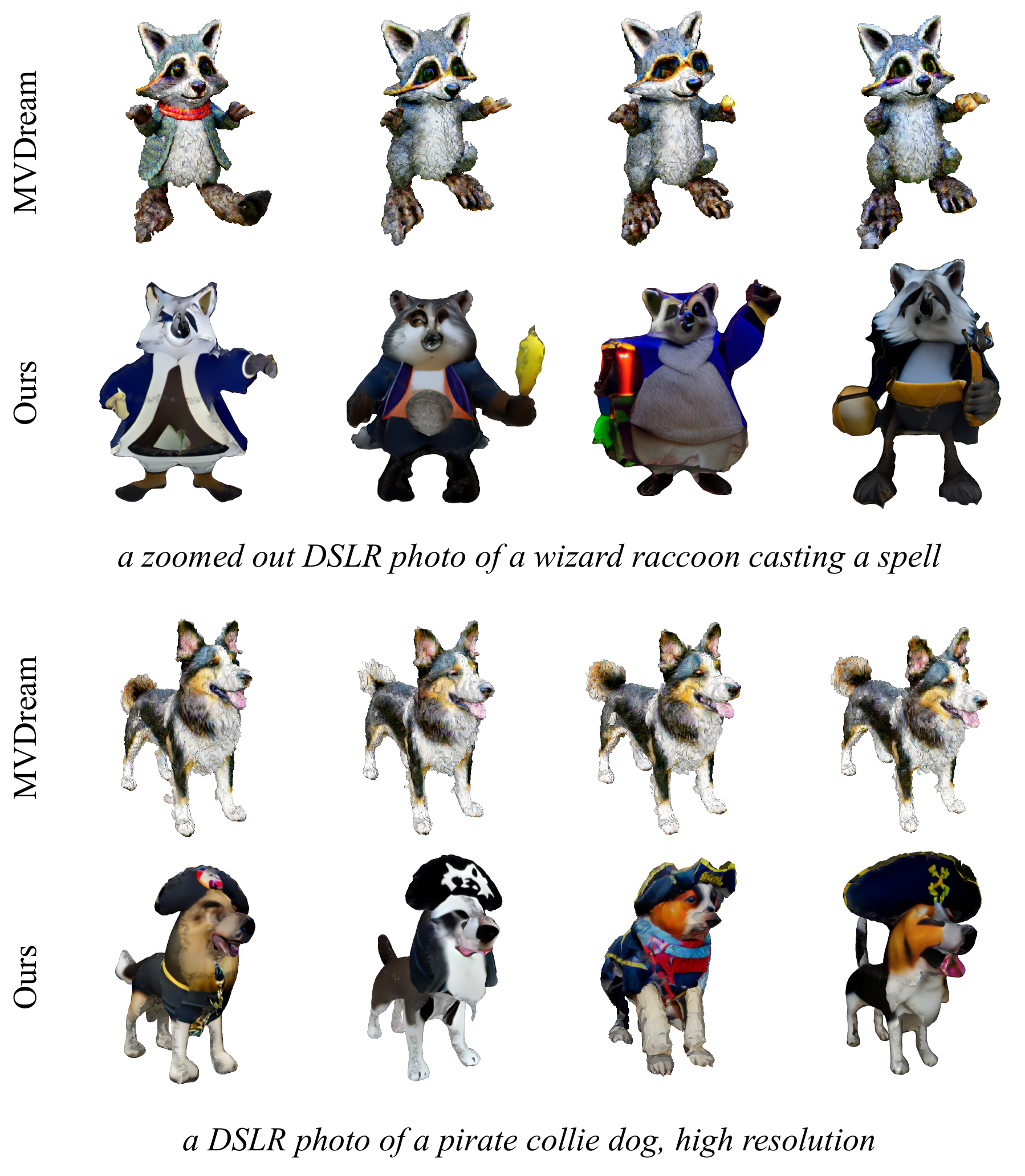}
    \vspace{-6mm}
    \caption{Comparison of sample diversity. Multiple samples are generated from the same prompt with different seeds. Our method is able to generate various samples while MVDream generates extremely similar results due to the SDS's mode-seeking nature. }
     \vspace{-5mm}
    \label{fig:var-comp}
\end{figure}

\section{Experiments}
In the following, we represent the experiment results of our approach and evaluate the design of our system, including qualitative comparisons against state-of-the-art techniques and quantitative evaluations of model performances. 

\subsection{Text-to-3D contents generation}
Given a random input text prompt, the proposed system is able to generate a high-fidelity 3D triangle mesh. Fig.~\ref{fig:results_gallery} shows a gallery of our generation results. Generated multi-view normal and RGB images are also presented beside the 3D mesh. Our multi-view normal diffusion model is able to generate high-quality normal maps with expressive geometry details, and the normal-conditioned RGB diffusion model also generates detailed textures aligned with input normal maps, which validates the effectiveness of our cross-view attention design. All prompts used are unseen during training, which proves the generalization ability.

\subsection{Qualitative and Quantitative Evaluation}
\fakepara{Qualitative evaluation}
In this section we compare our method with SDS-based methods including DreamFusion \cite{poole2022dreamfusion}, Fantasia3D \cite{chen2023fantasia3d}, and MVDream \cite{shi2023mvdream}. We also compare with the direct 3D generation methods including Point-E \cite{nichol2022point} and Shap-E \cite{jun2023shap}. The text prompts are provided from DreamFusion, which were unseen during the fine-tuning for MVDream and ours. 
Fig.~\ref{fig:qual-comp} illustrates qualitative comparisons of the renderings. It is clearly found that Point-E and Shap-E fail to generate reasonable text-aligned results. These direct 3D-based generation methods were trained on the relatively small 3D dataset compared to large-scale 2D text-image datasets, leading to poor generalization ability. Besides, DreamFusion and Fantasia3D suffer from the multi-face problem, while the results from the latter contain more details because of the supervision on geometry only. The rest two methods are 3D-aware so are able to produce reasonable 3D topology. MVDream generally achieves better visual quality, while our results are more consistent with the text prompts and take much less time to generate (35 mins v.s. 10s). 

\fakepara{Sample diversity}
Here, we compare the diversity of generated samples with MVDream. In this experiment, We generate 10 samples with the same prompt but different seeds. Fig.~\ref{fig:var-comp} presents the experiment results. Although both multi-view diffusion models are regularized by large-scale image-caption datasets to prevent overfitting on the 3D dataset, the results from MVDream still collapse to a single type because of the mode-seeking nature of SDS. On the contrary, our method can still keep the content diversity of the pre-trained diffusion model because the construction of 3D models is independent of the diffusion process, which would faithfully follow the random denoising process.

\begin{table}[t]
  \centering
  \resizebox{\linewidth}{!}{
  \begin{tabular}{@{}l||cccc@{}}
    \toprule
    {Settings \textbackslash \ Metrics} &  FID($\downarrow$) & IS ($\uparrow$) & CLIP ($\uparrow$)  \\
    \midrule
    \midrule
    Groundtruth normal renderings & $-$ & 9.17 & 0.279\\
    \quad (T2N) w/o 2D joint training & 43.61 & 8.94 & 0.270\\
    \quad (T2N) fewer 3D training data & 37.29 & 9.44 & 0.294\\
    \quad (T2N) proposed & 36.08 & 9.39 & 0.289\\
    \midrule
    \midrule
    Groundtruth RGB renderings & $-$ & 11.31 & 0.261 \\
    \quad (N2I) proposed & 35.40 & 11.25 & 0.257 \\
    \bottomrule
  \end{tabular}
  }
  \vspace{-2mm}
  \caption{We evaluate the proposed two multi-view diffusion models by computing. FID~\cite{heusel2017gans} (lower is better), IS ~\cite{salimans2016improved} (higher is better), and CLIP scores~\cite{radford2021learning} (higher is better) are used to measure the performance of different model variants. }
  \label{tab:quantitative}
  \vspace{-6mm}
\end{table}

\fakepara{Quantitative evaluation}
In the following, we quantitatively evaluate image generation quality and the text-image consistency of the proposed two novel multi-view diffusion models. Table \ref{tab:quantitative} demonstrates the evaluation results. Specifically, Frechet Inception Distance (FID)~\cite{heusel2017gans} and Inception Score (IS)~\cite{salimans2016improved} are adopted to measure the generation image quality and CLIP score cosine similarity \cite{radford2021learning} is calculated to measure the text-image consistency. We randomly select 2000 subjects as well as their multi-view RGB and normal renderings in the Objaverse \cite{deitke2023objaverse} dataset as our evaluation database. FID and IS are calculated independently of viewpoints while the CLIP similarity is selected as the max value across all 4-view scores. 

In general, we find that the proposed model achieves similar or even better results compared to the groundtruth renderings, which proves the high image quality and image-text consistency. We also evaluate the training strategies used in multi-view normal diffusion training, including using 2D large-scale dataset joint training, using higher consistency but fewer 3D subjects for training. It is clearly shown that the performance drastically drops when training without a 2D wild dataset injection. We believe that this is because fine-tuning purely multi-view normal data, would lead to a catastrophic forgetting of the original learned distribution and leads to poor learning ability. 
Training using fewer but higher text-consistent data leads to better IS and CLIP similarities, but worse FID. In practice, we found this model has lower generalization ability and diversity compared to the model that used more 3D data.

We also compare to previous SOTA methods quantitatively in Table \ref{tab:quantitative}. We randomly selected 50 prompts from Dreamfusion, not seen during our method and MVDream's fine-tuning, as the evaluation set. We adopt IS, CLIP scores and FID (Objaverse rendering and COCO validation set) to evaluate rendering results. Running time is also preseneted. Our method outperforms direct 3D diffusion methods significantly across all metrics and is on par with state-of-the-art SDS-based methods.
Our method achieves slightly better CLIP scores and FID but worse IS compared to MVDream, and consumes significantly less time for generation.

Please check the supplementary for more evaluations.

\begin{table}[t]
  \centering
  \resizebox{\linewidth}{!}{
  \begin{tabular}{@{}l||cccccc@{}}
    \toprule
    {Methods \textbackslash \ Metrics}  & IS ($\uparrow$) & CLIP ($\uparrow$) & FID ($\downarrow$ objv.) & FID ($\downarrow$ COCO) & Run Time \\
    \midrule
    Point-E  & 7.265 & 0.220 & 104.105 & 164.765 & $\sim 20$ s\\
    Shap-E  & 7.412 & 0.236 & 103.557 & 163.105 & $\sim 4$ s\\
    \midrule
    Dreamfusion  & 7.724 & 0.245 & 125.873 & 150.285 & $\sim 50$ m\\
    Fantasia3d  & 8.311 & 0.207 & 132.941 & 150.255 & $\sim 115$ m\\
    ProlificDreamer  & 9.457 & 0.269 & 121.577 & 124.185 & $\sim 5$ h\\
    MVDream  & 8.180 & 0.262 & 117.715 & 133.089 & $\sim 35$ m\\
    \midrule
    Ours  & 8.111 & 0.267 & 82.324 & 126.014 & $\sim 10$ s\\
    \bottomrule
  \end{tabular}
  }
  \vspace{-3mm}
  \caption{Quantitative comparisons with previous methods. }
  \label{tab:quantitative}
   \vspace{-5mm}
  
\end{table}

\section{Limitations and Future Work}

\fakepara{Limited view numbers}
Due to the small view number, areas such as top, bottom and concavity cannot be fully observed, and thus their geometry or appearance cannot be well reconstructed. Apart from the iterative update scheme, the multi-view diffusion can be extended to more views.

\fakepara{Texture quality}
For the appearance, we finetune a multi-view normal-conditioned diffusion model for efficiency. However, the ability of generating realistic images is degraded due to the texture quality of the 3D training samples and their rendering quality. Apart from further enhancing the training samples, we can also apply the state-of-the-art texture generation systems \cite{chen2023text2tex} for non-time-sensitive tasks. 

Please check the supplementary for more discussions.

\section{Conclusion}
We propose to perform fast text-to-3D generation by fine-tuning a multi-view 2.5D diffusion from pre-trained RGB diffusion models. To learn multi-view consistency, the model is fine-tuned on multi-view normal map renderings, with cross-view attention as the structural guarantee. After the simultaneous generation of multi-view normal maps, 3D models are obtained by deforming meshes by differentiable rasterization. Finally, appearance is generated by multi-view normal-conditioned RGB diffusion. Our generation pipeline produces diverse and high-quality 3D models in 10 seconds, and demonstrates strong generalization to complex content and generates fine details. Extensive experiments are conducted to show that our method enables fast generation of realistic, complex, and diverse models.

\section{Acknowledgments}
This work is supported by National Natural Science Foundation of China under Grants 62001213 and Hong Kong RGC GRF 16206722.

{
    \small
    \bibliographystyle{ieeenat_fullname}
    \bibliography{main}
}

\clearpage

\twocolumn[
\begin{center}
	\textbf{\LARGE ------ Supplementary Material ------}
	\vspace{5mm}
\end{center}
]
\setcounter{section}{0}
\setcounter{equation}{0}
\setcounter{figure}{0}
\setcounter{table}{0}
\setcounter{page}{1}
\makeatletter

Due to the space limitation of the main paper, we provide supplementary materials to give an auxiliary demonstration. In this PDF file, we will present a detailed description of the implementation details, additional evaluation and discussions, and more results. We also provide a project page to present video results for better visualization. Project page: \href{https://nju-3dv.github.io/projects/direct25}{https://nju-3dv.github.io/projects/direct25}.

\section{Implementation Details}
In this section, we describe more implementation details of the proposed system, including data preparation, iterative updating, inference time, and another texturing implementation.

\subsection{Dataset Preparation}
\label{sec:4.1}
We use the Objaverse~\cite{deitke2023objaverse} dataset for 2.5D training data generation, which is a large-scale 3D object dataset containing 800K high-quality models. We use the captions provided by Cap3d \cite{luo2023scalable} as text prompts, which is the best 3D dataset caption method currently. Each object is firstly normalized at the center within a bounding box $[-1, 1]^3$, and we render the scene from 32 viewpoints uniformly distributed in azimuth angles between $[0^\circ, 360^\circ]$. The elevation is set to $0^\circ$ and camera FoV is set to $60^\circ$. The camera distance from the origin $(0, 0, 0)$ is set to a fixed distance equal to 1.5 times the focal length in normalized device coordinates. For lighting, we use a composition of random lighting selected from point lighting, sun lighting, spot lighting, and area lighting. RGB images and normal maps in world coordinates are rendered using a rasterizer-based renderer for each object.

Besides, we also adopt a large-scale 2D image-text dataset to improve the generation diversity following mvdream~\cite{shi2023mvdream}. Specifically, we use the COYO-700M dataset~\cite{kakaobrain2022coyo-700m}, which also contains metadata like resolution and CLIP scores~\cite{radford2021learning}, etc. We filter the dataset with both width and height greater than 512, aesthetic scores \cite{schuhmann2022improved} greater than 5, and watermark scores lower than 0.5, which results in a 65M-size subset. Though the filtered dataset is reduced to $1/10$ of the original size, it is still much larger than the 3D dataset. Actually, we do not consume the whole dataset within the designated training time. In the following, we describe the specific dataset usage for two proposed multi-view diffusion model training.

\noindent \textbf{Text-to-normal multi-view diffusion model.} As we want to generate high-quality and multi-view consistent normal maps from a single text prompt input, we are able to use all valid normal map renderings in Objaverse~\cite{deitke2023objaverse}. We filter the dataset by sorting the CLIP similarities between RGB images and captions and selecting the top 500K objects to keep a high text-image consistency. We take a similar 2D \& 3D joint training strategy with MVDream~\cite{shi2023mvdream}, where 3D data and 2D data are randomly chosen in each batch with a probability of 80\% and 20\%, respectively. This trick can guarantee the same expected number of instances to be seen in each training step because 4 views are from the same object for 3D dataset. Also for 3D data, we add a special tag \textit{normal map} to the end of captions to indicate the normal map prediction task. During inference, we also add this postfix to the prompt for normal map predictions.

\noindent \textbf{Normal conditioned RGB multi-view diffusion model.} Some samples in the Objaverse dataset has cartoonish appearance, and we would like to filter out these samples. Specifically, we first filter the dataset to obtain renderings whose aesthetic scores are larger than 5, which results in a 130K subset. Then, we compute the CLIP scores between the remaining images and two pre-defined positive and negative quality description prompts \footnote{Positive prompt: realistic, 4K, vivid, highly detailed, high resolution, high quality, photography, HD, HQ, full color; 
Negative prompt: cartoon, flat color, simple texture, ugly, dark, bad anatomy, blurry, pixelated obscure, unnatural colors, poor lighting, dull, unclear, cropped, lowres, low quality, artifacts, duplicate, morbid, mutilated, poorly drawn face, deformed, dehydrated, bad proportions}. We compute the ratio of the positive scores and negative scores and select the top 10K data as our training dataset. We found that this strategy successfully selected the high-quality renderings in the dataset, and works better than training on all rendering data.

\subsection{Iterative Updating}
In most cases, a single run of the pipeline is enough to generate high-quality results. However, for some topologies, there may be large areas unobserved by the 4 perspectives (e.g., large planar areas on the top of the object). To address this issue, we could iteratively update rendered images from novel views by the inpainting \cite{lugmayr2022repaint} pipeline to refine the texture. Specifically, we compute an inpainting mask indicating the unseen areas at a new camera viewpoint, and the invisible areas are edited given a certain noise strength. In Fig.~\ref{fig:iterative}, we present the results of the iterative updating. In this example, we inpaint the top views of the generated bread and fuse the resulted RGB images back to the generated model. As shown in the figure, the top areas of the bread are unseen during the first generation, and we inpaint the unseen areas in the second run. The inpainting mask is used to ensure that only the unseen areas would be modified, while other regions are kept unchanged. The final generated model (Fig \ref{fig:iterative} (e)) demonstrates the effectiveness of the strategy. During experiments, we found that 1 or 2 iterations suffice to recover the unseen areas.

\subsection{Inference Time}
Compared to SDS optimization-based methods which typically take over half an hour, our method is efficient enough to generate high-quality results in 10 seconds: On a single Nvidia A100 GPU, the denoising process of the two multi-view diffusion models each takes around 2.5 seconds for 50 DDIM steps. The explicit geometry optimization takes around 2 $\sim$ 3 seconds for 200 optimization steps, which depends on the triangle mesh complexity. The final texture fusion takes around 1.5 seconds. the efficiency and diversity of the proposed system enable selection from batch generated samples, which greatly increases the practicality for prototyping and digital content creation. For iterative updating, typically 1-3 passes are enough to paint the unseen areas and can be finished in less than one minute, which is still much faster than the previous SDS optimization-based methods.

\begin{table}[t]
  \centering
  \resizebox{\linewidth}{!}{
  \begin{tabular}{@{}c||cccccc@{}}
    \toprule
    {views \textbackslash \ steps}  & 50 & 100 & 200 & 400 & 600 \\
    \midrule
    4-view  & 0.78 / 1.57 & 0.81 / 1.42 & 0.82 / 1.11 & 0.82 / 1.20 & 0.81 / 1.22\\
    8-view  & 0.81 / 1.20 & 0.85 / 1.18 & 0.85 / 0.99 & 0.84 / 1.15 & 0.85 / 1.21 \\
    16-view  & 0.82 / 1.14 & 0.85 / 0.91 & 0.86 / 1.06 & 0.86 / 1.07 & 0.86 / 1.06 \\
    \bottomrule
  \end{tabular}
  }
  \vspace{-2mm}
  \caption{Normal consistency ($\uparrow$) and Chamfer-Distance ($\downarrow $) evaluation for fast meshing under view and optimization step numbers. }
  \label{tab:fast_meshing}
  
\end{table}

\begin{figure}[t]
  \centering
  \includegraphics[width=\linewidth]{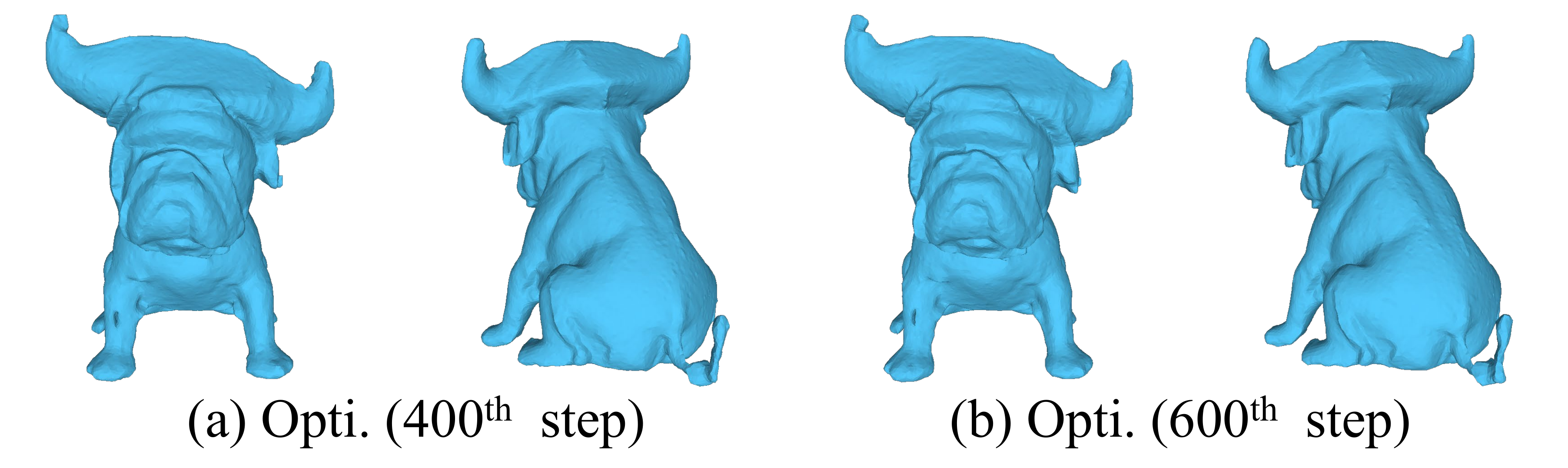}
   \vspace{-6mm}
   \caption{Visualization of more than 200 optimization steps.}
   \label{fig:fig1}
   \vspace{-4mm}
\end{figure}

\subsection{Alternative Texturing Implementation}
Besides the mentioned texturing method in the main paper, we also propose an alternative optimization-based texturing method as our open-source version. Similar to geometry optimization, we optimize the texture map in UV space by minimizing the reconstruction loss of the multi-view RGB images. Specifically, we adopt the $L_1$ RGB loss, SSIM loss, and a total variation (TV) loss on the UV texture map as a regularization. The weights for these three losses are set to 1.0, 10.0, and 1.0. In experiments, we found that only $50 - 100$ steps are enough to generate satisfactory results, and the optimization takes only about 1 second.

\begin{figure*}[htbp]
    \centering
    \includegraphics[width=\linewidth]{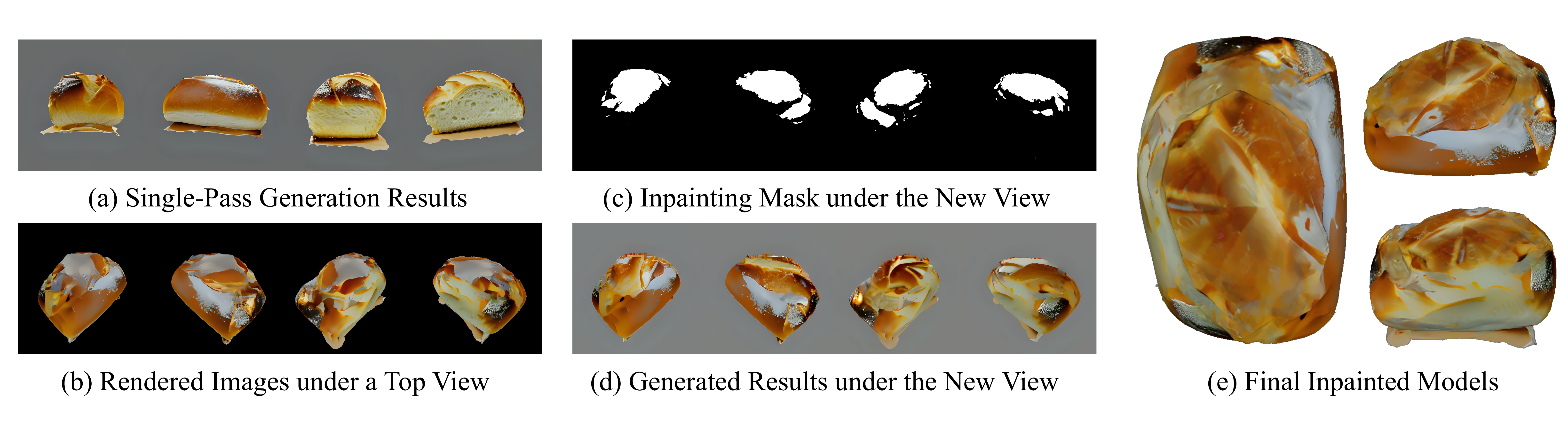}
    \caption{Demonstration of the iterative updating. (a) is the single-pass generated multi-view RGB images given a prompt "a freshly baked loaf of sourdough bread on a cutting board”. (b) shows the rendered results of the single-pass generated model. As seen, the top area remains uncolored. (c) shows the generated inpainting mask under the new view, where the white areas denote the areas that are invisible and need to be inpainted. (d) is the inpainted results under the new view given the previously rendered results and the visibility mask. (e) demonstrates the final generated mesh under the top view and two side-top views. The previous uncolored areas now have been inpainted with reasonable and coherent colors.}
    \label{fig:iterative}
\end{figure*}

\begin{figure*}[htbp]
    \centering
    \includegraphics[width=\linewidth]{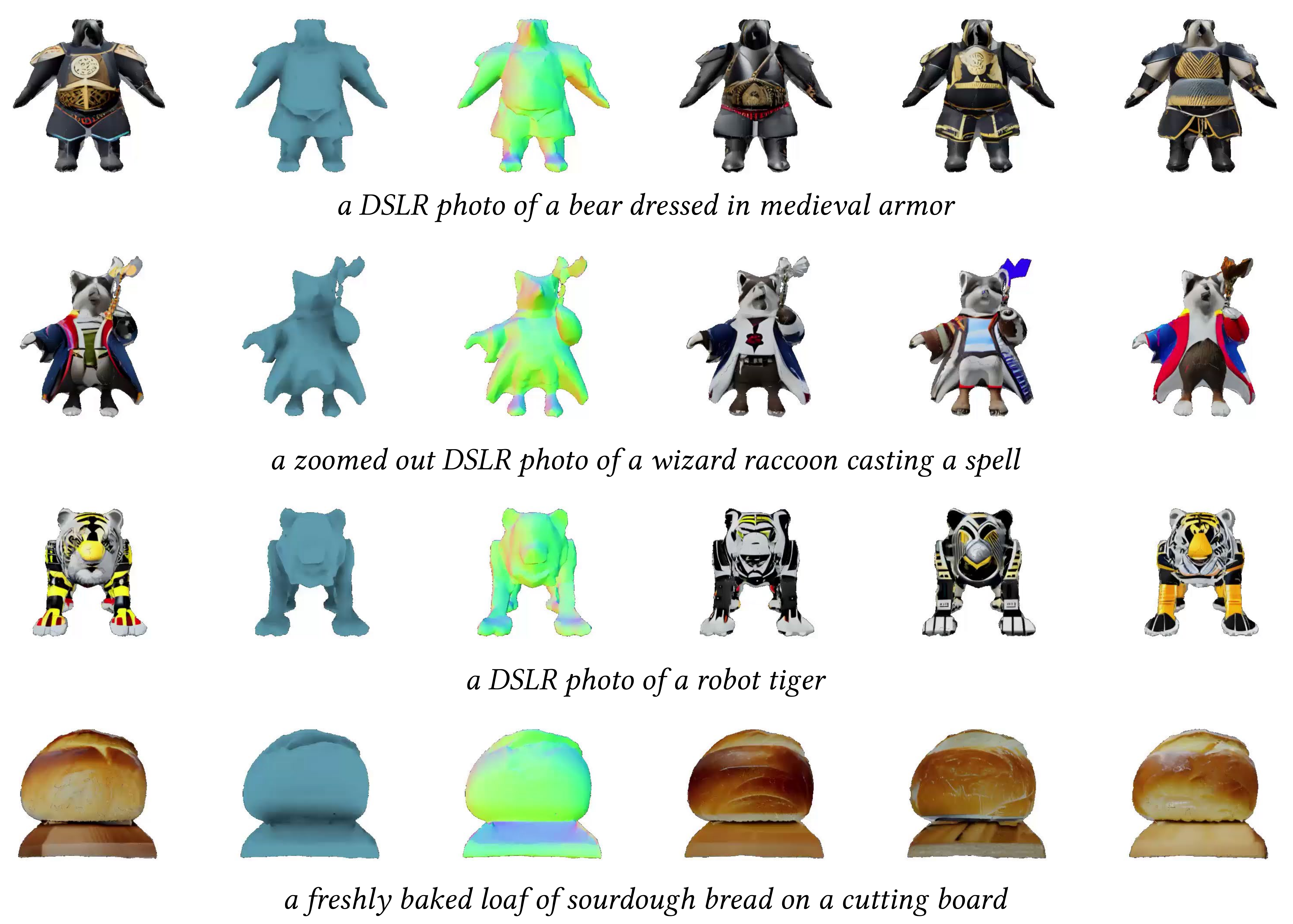}
    \caption{Demonstration of Geometry-appearance disentangled generation. Due to the two-stage sequential setting, our method greatly increases the control ability of the content generation results.}
    \label{fig:disentangled}
\end{figure*}

\section{Geometry-Appearance Disentangled Generation}
\label{sec:disentangle}
Due to the two-stage setting in the proposed method, one could generate random RGB images while keeping the geometry fixed, which enables geometry-appearance disentangled generation and offers better control over the generation process. Fig.~\ref{fig:iterative} demonstrates the disentangled generation results. It demonstrates that users can fix the satisfying generated geometry and then proceed to appearance generation. 

\section{More Evaluations}
Here we present more evaluations of the proposed fast meshing algorithm. Specifically, We present normal consistency and Chamfer Distance ($\times 10^{-3}$) evaluation w.r.t.~optimization steps (\cref{tab:fast_meshing}) of the fast meshing on 15 chosen Objaverse meshes with ground truth normal maps. In most cases, we found no obvious improvements beyond 200 steps, as also shown in \cref{fig:fig1}.

The number of views for reconstruction is constrained by the diffusion model: SD2.1 512x512 resolution aligns with four 256x256 views. Our early experiments suggested optimal quality when the total pixel count aligns with the base model's resolution. So we also do this ablation study by feeding ground truth to the optimization like the previous one. We find that more views lead to slightly better reconstruction, and leave this for future work. 

\section{Discussions}
In the following, we provide a detailed discussion about the settings of our system, including the two-stage sequential models, and normal predictions v.s.~depth predictions.

\noindent \textbf{Two-stage sequential architecture}. As demonstrated in Sec.~\ref{sec:disentangle}, a two-stage sequential architecture naturally enables the geometry-appearance disentangled generation and provides more freedom on both geometry and appearance generation. Besides, using a combined pipeline also leads to a double GPU memory requirement compared to the sequential setting, which could become a great burden under the multi-view setting. This challenge becomes much more severe when one increases the spatial resolution of the diffusion model, e.g.~from 256 to 512 or even 1024. Finally, the sequential model has better multi-view and geometry-appearance consistency. Instead of the generation normal maps, we use the ones rendered from the optimized mesh for the texture diffusion model input. On the one hand, the rendered normal maps are guaranteed to be consistent. On the other hand, it provides better alignment between the generated RGB images and the actual geometry. For the above reasons, our system takes the two-stage sequential as our architecture.

\noindent  \textbf{Normal v.s.~Depth}. Another alternative choice for our system is to use depth instead of normal. Because normal is the first-order derivative of the depth, it is free from scale ambiguity and provides a higher tolerance for multi-view inconsistency. Optimizing depth value directly requires much higher multi-view accuracy and therefore decreases the robustness of the geometry optimization system. Previous work~\cite{wang2022neuris} also found that using normal priors performs better than the depth priors, which also supports our assumption. 
Secondly, normal serves as a better conditioning signal for RGB generation because it generally has better alignment than depth. For example, sharp normal changes result in RGB discontinuity because of shading, but in this case depth may still be smooth. 
Therefore, we adopted normal as our shape representations and found it worked well.

\section{Limitations and Failure Cases}
In the main paper, we briefly discuss the limitations of the proposed pipeline and here we present more discussions.

\fakepara{Multi-view consistency}
The multi-view RGB/normals are generated by the self-attention mechanism in the multi-view diffusion models without any physical-based supervision, which means the multi-view consistency is not guaranteed. This is an inherent issue in multiview diffusion models like MVDream, which is known to be prone to geometric misalignment. Our first stage text-to-normal diffusion model also suffers from this issue, while we found that the issue on the normal model is smaller than that on the RGB model. Besides, the second stage adopts multiview-consistent rendered normals as input, which relieves challenges faced by one-stage models like MVDream. We will include this part in future work.

\begin{figure}[t]
  \centering
  \includegraphics[width=\linewidth]{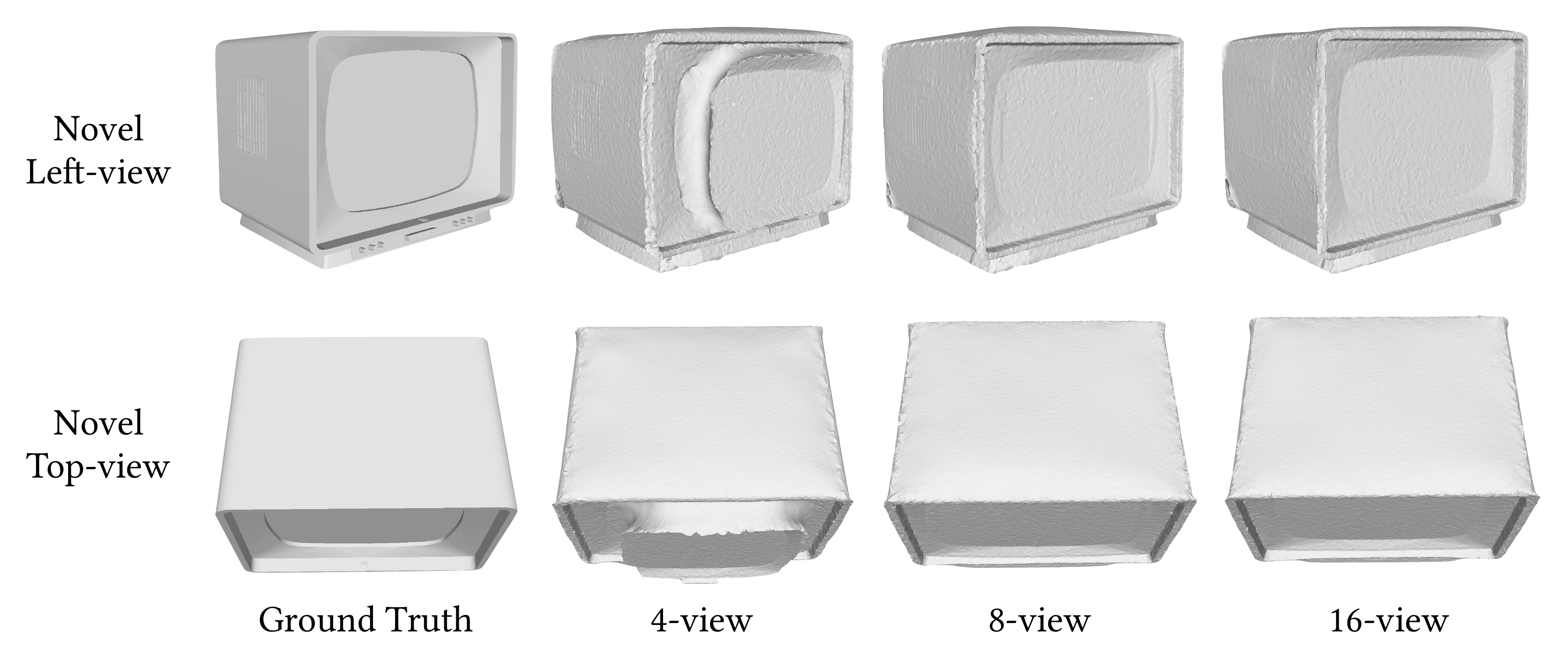}
   \vspace{-6mm}
   \caption{Failure case. The normal-based reconstruction system suffered from the depth ambiguity issue. In this example, 4-view reconstruction fails on the television screen and introduces artifacts. Using more views solves this problem. }
   \label{fig:failure}
\end{figure}

\fakepara{Limited view numbers and failure cases}
Because the number of views is small, areas such as top, bottom, and concavity cannot be fully observed, and thus their geometry or appearance cannot be well reconstructed. Apart from the iterative update scheme, the multi-view diffusion model can be further extended to handle more views. 

We also emphasize that the limited view of normal maps may not provide sufficient information for reconstruction, leading to a degraded performance as shown in \cref{tab:fast_meshing}. We also present an example in \cref{fig:failure}. This is due to the intrinsic issue brought about by the normals being second-order derivatives of world positions, which introduces ambiguities in shape, i.e., we only know its direction in the world coordinate system, but not its specific depth, because the normal maps are the same for any depth. As shown in \cref{fig:failure}, the television screen failed to reconstruct given only 4 views normals. The full screen overall protrudes, but under the training viewpoints, there is no difference in normals - they are pointing in the same direction. This is an inherent issue with reconstruction based on normals, and using more viewpoints can greatly alleviate this problem. Our reconstruction system also suffers from this issue, and more views could lead to more accurate reconstruction.

\fakepara{Texture quality}
For the appearance, we finetune a multi-view normal-conditioned diffusion model for efficiency. However, the ability to generate realistic images is degraded due to the texture quality of the 3D training samples and their rendering quality. Apart from further enhancing the training samples, we can also apply the state-of-the-art texture generation systems \cite{chen2023text2tex} for non-time-sensitive tasks.

\section{More Results}
\begin{figure*}[t]
    \centering
    \includegraphics[width=\linewidth]{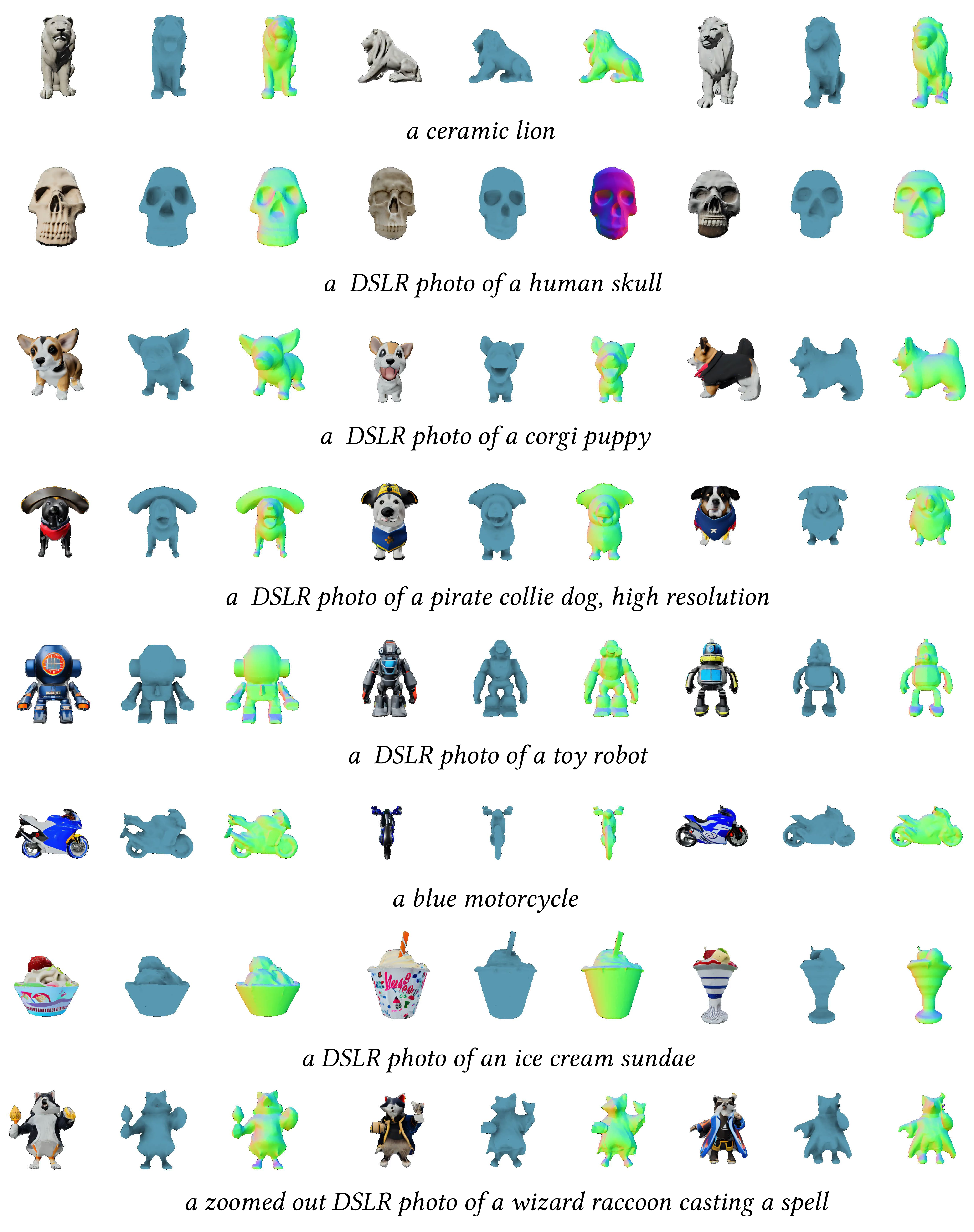}
    \caption{More Diverse Generation Results. Our method avoids the common mode-seeking problem by SDS and generates diverse results.}
    \label{fig:diversity}
\end{figure*}

\begin{figure*}[t]
    \centering
    \includegraphics[width=\linewidth]{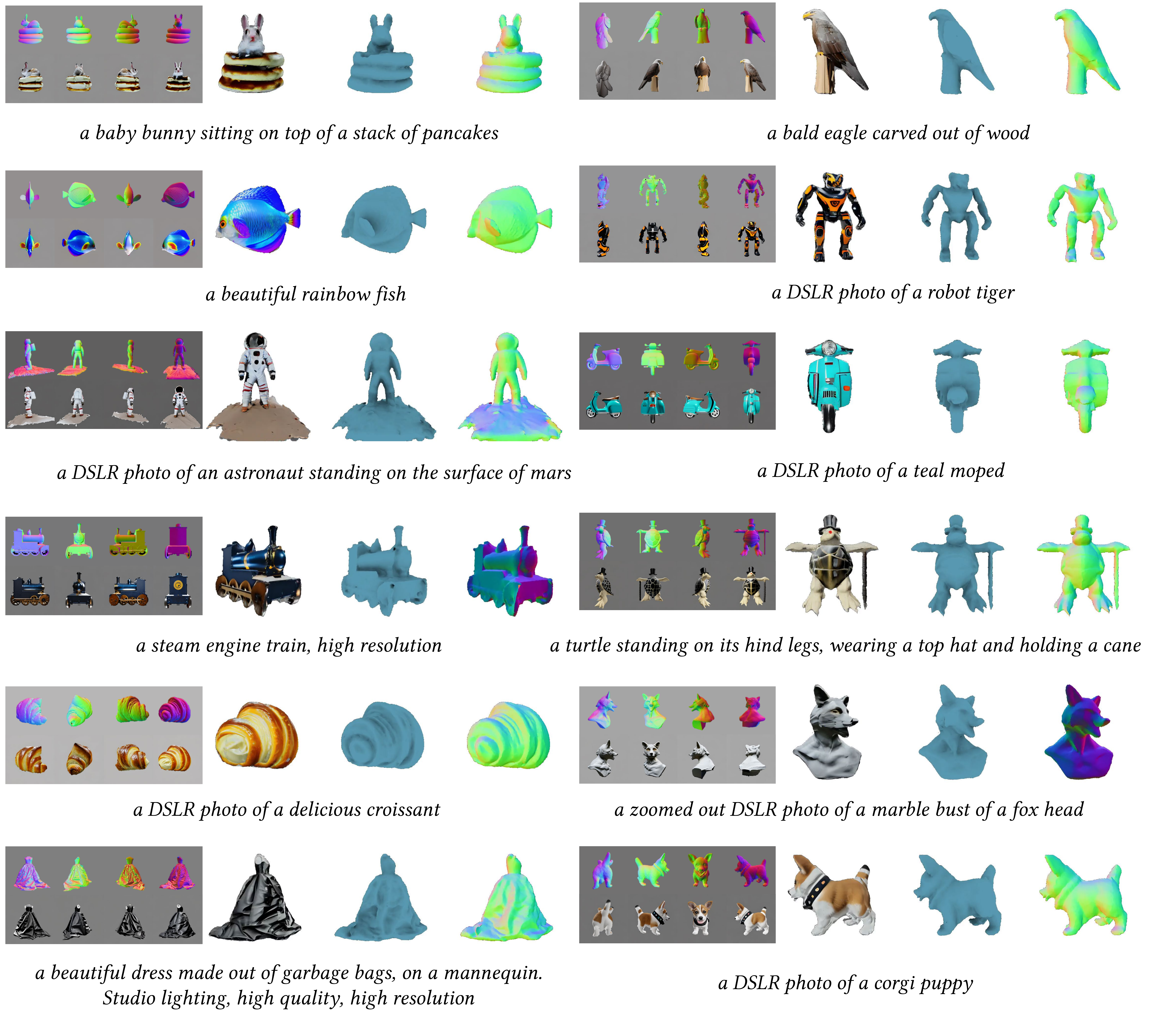}
    \caption{Results Gallery. Given text prompts as description input, our method outputs high-quality textured triangle mesh in only 10 seconds. The generated multi-view normal and RGB images are shown beside the rendered models. Prompts for the above left column results are R1) a baby bunny sitting on top of a stack of pancakes, R2) a beautiful rainbow fish, R3) a DSLR photo of an astronaut standing on the surface of mars, R4) a steam engine train, high resolution, R5) a DSLR photo of a delicious croissant, and R6) a beautiful dress made out of garbage bags, on a mannequin. Studio lighting, high quality, high resolution. Prompts for the above right column results are R1) a bald eagle carved out of wood, R2) a DSLR photo of a robot tiger, R3) a DSLR photo of a teal moped, R4) a turtle standing on its hind legs, wearing a top hat and holding a cane, R5) a zoomed out DSLR photo of a marble bust of a fox head, and R6) a DSLR photo of a corgi puppy.}
    \label{fig:gallery1}
\end{figure*}

\begin{figure*}[t]
    \centering
    \includegraphics[width=\linewidth]{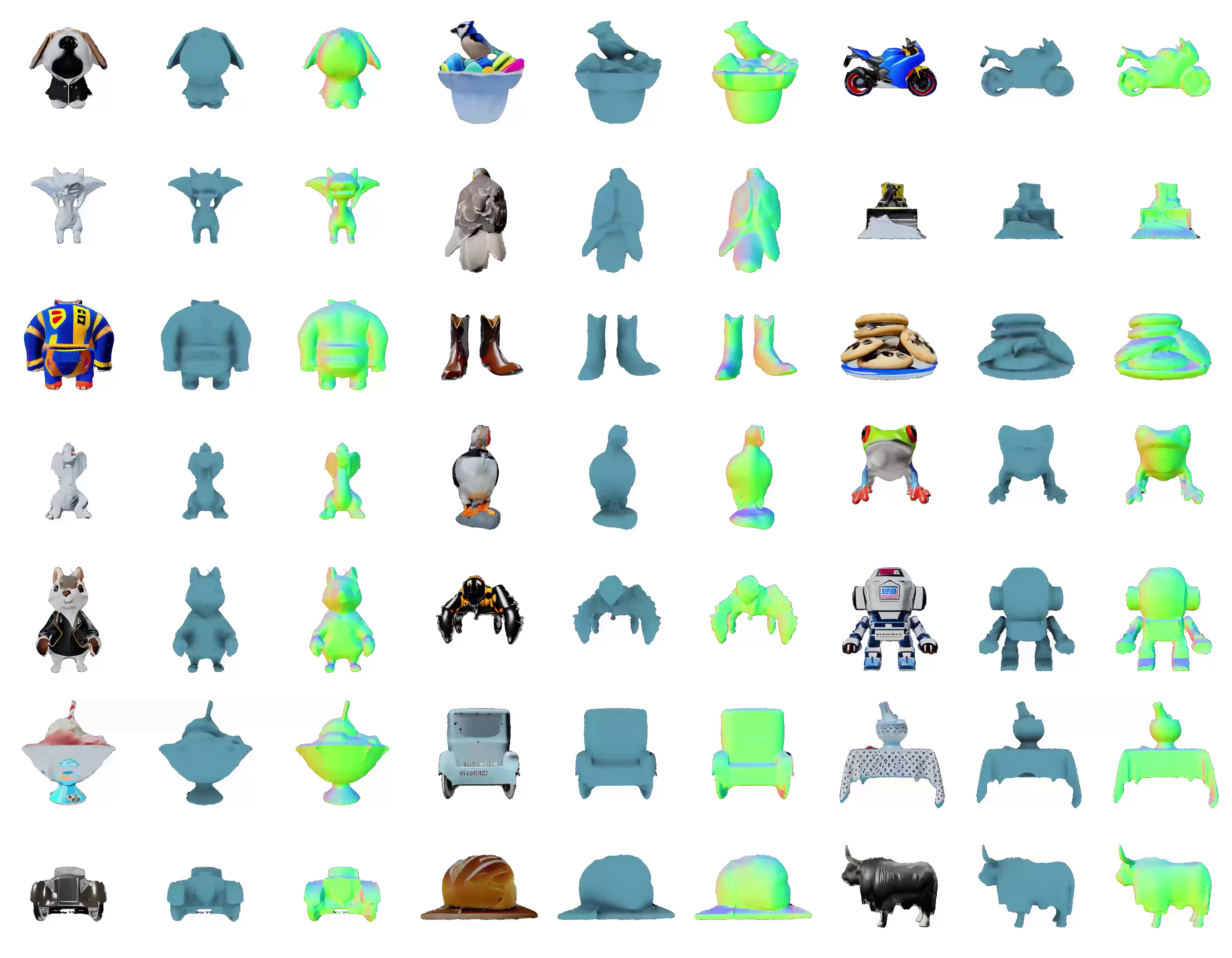}
    \caption{More generation results. Prompts for the above results from top to bottom and left to right are R1-l) a beagle in a detective's outfit, R1-2) a blue jay standing on a large basket of rainbow macarons, R1-3) a blue motorcycle, R2-1) a dragon-cat hybrid, R2-2) a DSLR photo of a bald eagle, R2-3) a DSLR photo of a bulldozer, R3-1) a DSLR photo of a hippo wearing a sweater, R3-2) a DSLR photo of a pair of tan cowboy boots, studio lighting, product photography, R3-3) a DSLR photo of a plate piled high with chocolate chip cookies, R4-1) a DSLR photo of a porcelain dragon, R4-2) a DSLR photo of a puffin standing on a rock, R4-3) a DSLR photo of a red-eyed tree frog, R5-1) a DSLR photo of a squirrel wearing a leather jacket, R5-2) a DSLR photo of a tarantula, highly detailed, R5-3) a DSLR photo of a toy robot, R6-1) a DSLR photo of an ice cream sundae, R6-2) an old vintage car, R6-3) a DSLR photo of an ornate silver gravy boat sitting on a patterned tablecloth, R7-1) a frazer nash super sport car, R7-2) a freshly baked loaf of sourdough bread on a cutting board and R7-3) a highland cow.}
    \label{fig:gallery2}
\end{figure*}

\begin{figure*}[t]
    \centering
    \includegraphics[width=\linewidth]{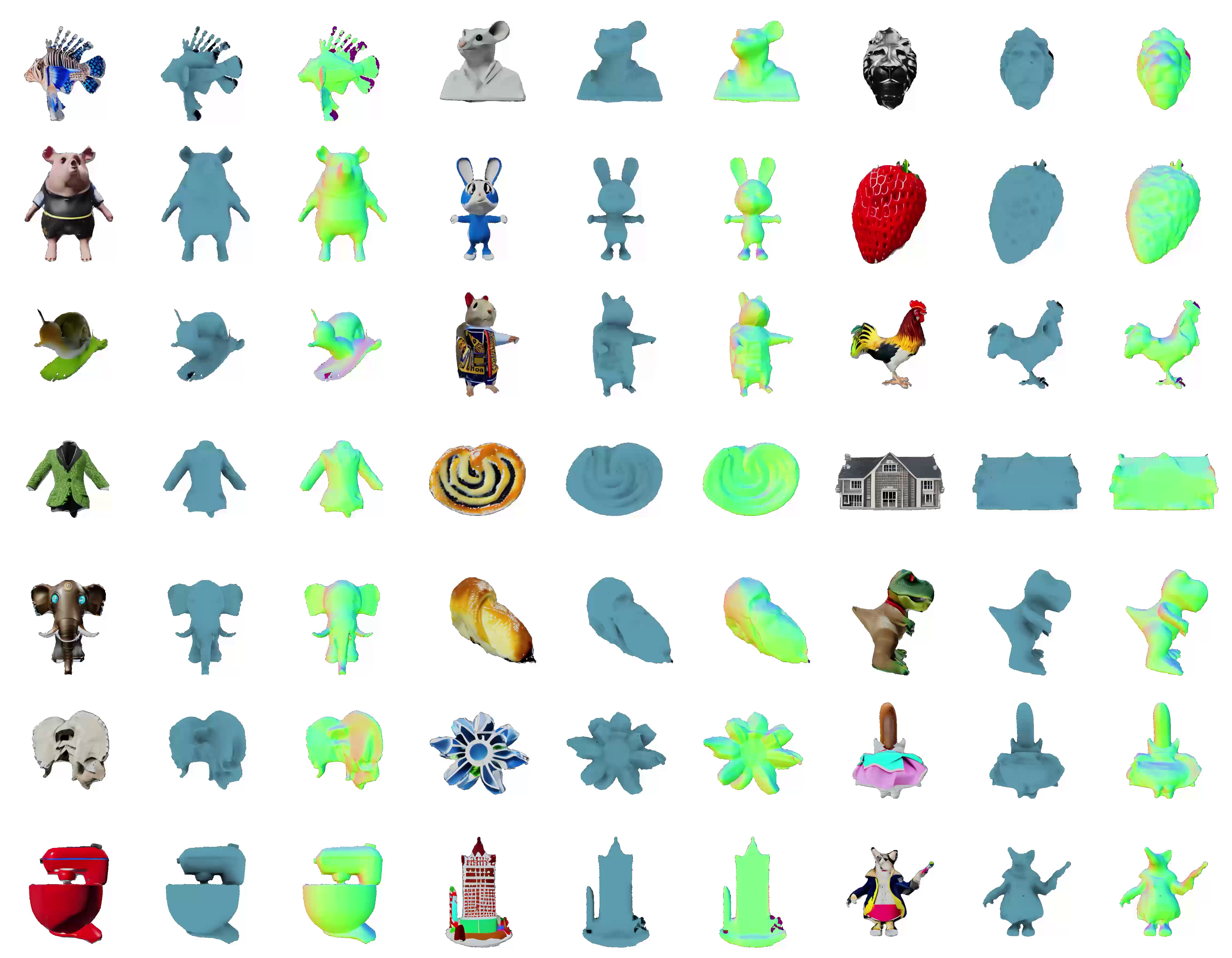}
    \caption{More generation results. Prompts for the above results from top to bottom and left to right are R1-l) a lionfish, R1-2) a marble bust of a mouse, R1-3) a metal sculpture of a lion's head, highly detailed, R2-1) a pig wearing a backpack, R2-2) a rabbit, animated movie character, high detail 3d model, R2-3) a ripe strawberry, R3-1) a snail on a leaf, R3-2) a squirrel dressed like Henry VIII king of England, R3-3) a wide angle DSLR photo of a colorful rooster, R4-1) a zoomed out DSLR photo of a beautiful suit made out of moss, on a mannequin. Studio lighting, high quality, high resolution, R4-2) a zoomed out DSLR photo of a fresh cinnamon roll covered in glaze, R4-3) a zoomed out DSLR photo of a model of a house in Tudor style, R5-1) a cute steampunk elephant, R5-2) a DSLR photo of a delicious croissant, R5-3) a DSLR photo of a plush t-rex dinosaur toy, studio lighting, high resolution, R6-1) a DSLR photo of an elephant skull, R6-2) a flower made out of metal, R6-3) a hotdog in a tutu skirt, R7-1) a shiny red stand mixer, R7-2) a wide angle zoomed out view of Tower Bridge made out of gingerbread and candy and R7-3) a zoomed out DSLR photo of a wizard raccoon casting a spell.}
    \label{fig:gallery3}
\end{figure*}

We present more results of the proposed method on the following pages, including the various generation ability (Fig \ref{fig:diversity}) and more generation results (Fig.~\ref{fig:gallery1}, \ref{fig:gallery2}, \ref{fig:gallery3}).

\section{Additional Video Results}
We present video results of the proposed method in our project page: \href{https://nju-3dv.github.io/projects/direct25}{https://nju-3dv.github.io/projects/direct25}. 

Please check it for better visualization.

\end{document}